\documentclass[conference]{IEEEtran}
\usepackage{times}

\usepackage[numbers]{natbib}
\usepackage{multicol}

\usepackage[utf8]{inputenc} %
\usepackage[T1]{fontenc}    %
\usepackage{booktabs}       %
\usepackage{amsfonts}       %
\usepackage{nicefrac}       %
\usepackage{microtype}      %

\usepackage[backref=page]{hyperref}

\usepackage[numbers]{natbib}

\usepackage{amsmath,amssymb}
\usepackage{graphicx}
\usepackage{bbm}
\usepackage{subcaption}
\usepackage[font=small]{caption}

\let\labelindent\relax
\usepackage{enumitem}
\usepackage{algorithm}
\usepackage{algpseudocode}
\usepackage{colortbl}
\usepackage{xcolor}
\usepackage{multirow}

\usepackage{titletoc}
\usepackage[T1]{fontenc}    %
\usepackage{url}            %

\usepackage{cleveref}

\usepackage{bbm}

\usepackage{siunitx}
\usepackage{tablefootnote}
\usepackage{makecell}
\usepackage{dsfont}

\IEEEoverridecommandlockouts                              %

\definecolor{mydarkblue}{rgb}{0,0.08,0.45}
\hypersetup{ %
    colorlinks=true,
    linkcolor=mydarkblue,
    citecolor=mydarkblue,
    filecolor=mydarkblue,
    urlcolor=mydarkblue,
}

\newcommand{\ie}{\textit{i}.\textit{e}.\ }
\newcommand{\eg}{\textit{e}.\textit{g}.\ }

\newcommand{\mytable}[1]{Table~\ref{#1}}

\newcommand{\mysec}[1]{Section~\ref{#1}}
\newcommand\mypara[1]{\vspace{1mm}\noindent\textbf{#1}}

\def\1{\mathds{1}}

\begin{document}

\title{DexPBT: Scaling up Dexterous Manipulation for \\ Hand-Arm Systems with Population Based Training}

\author{Aleksei Petrenko, Arthur Allshire, Gavriel State, Ankur Handa, Viktor Makoviychuk}

\makeatletter
    \let\@oldmaketitle\@maketitle%
    \renewcommand{\@maketitle}{\@oldmaketitle
    \begin{minipage}[c]{\textwidth}
    \centering
    \vspace{2pt}
    \includegraphics[width=0.245\linewidth]{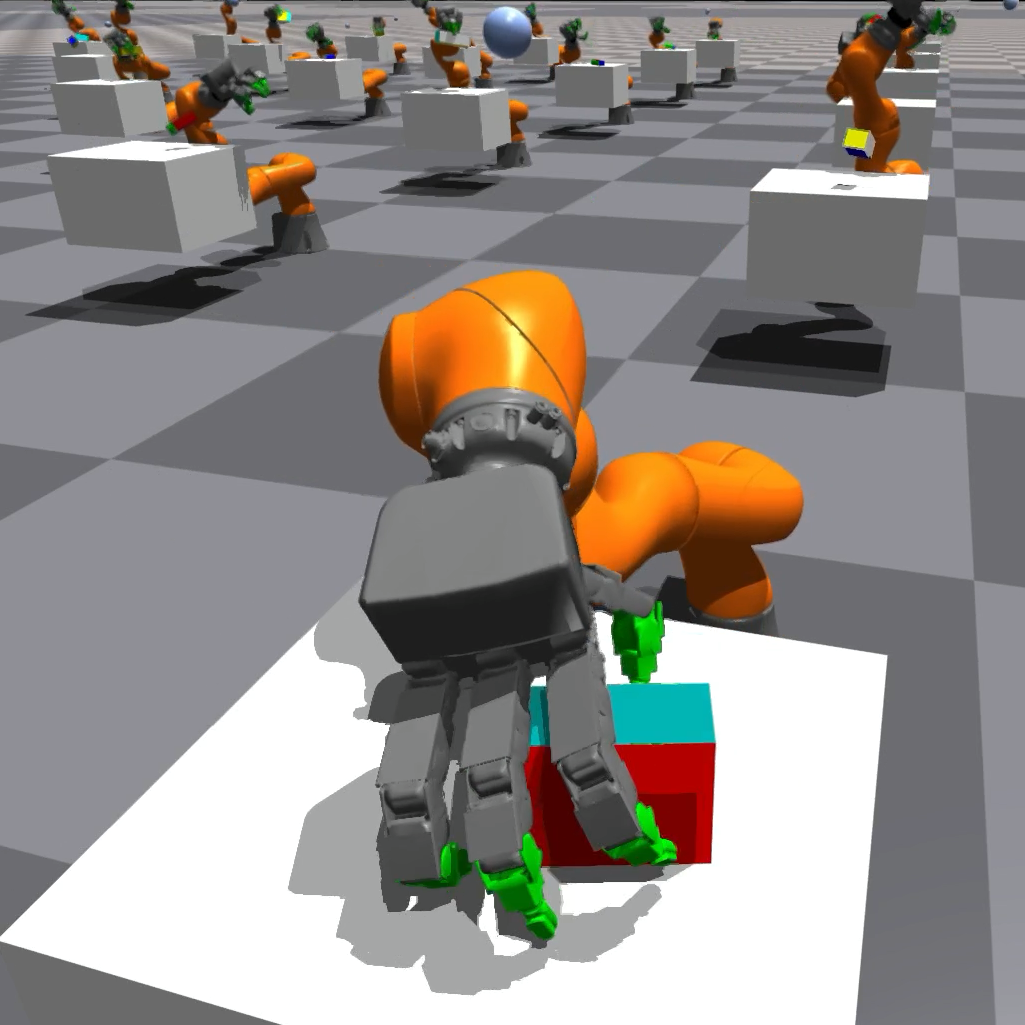}
    \includegraphics[width=0.245\linewidth]{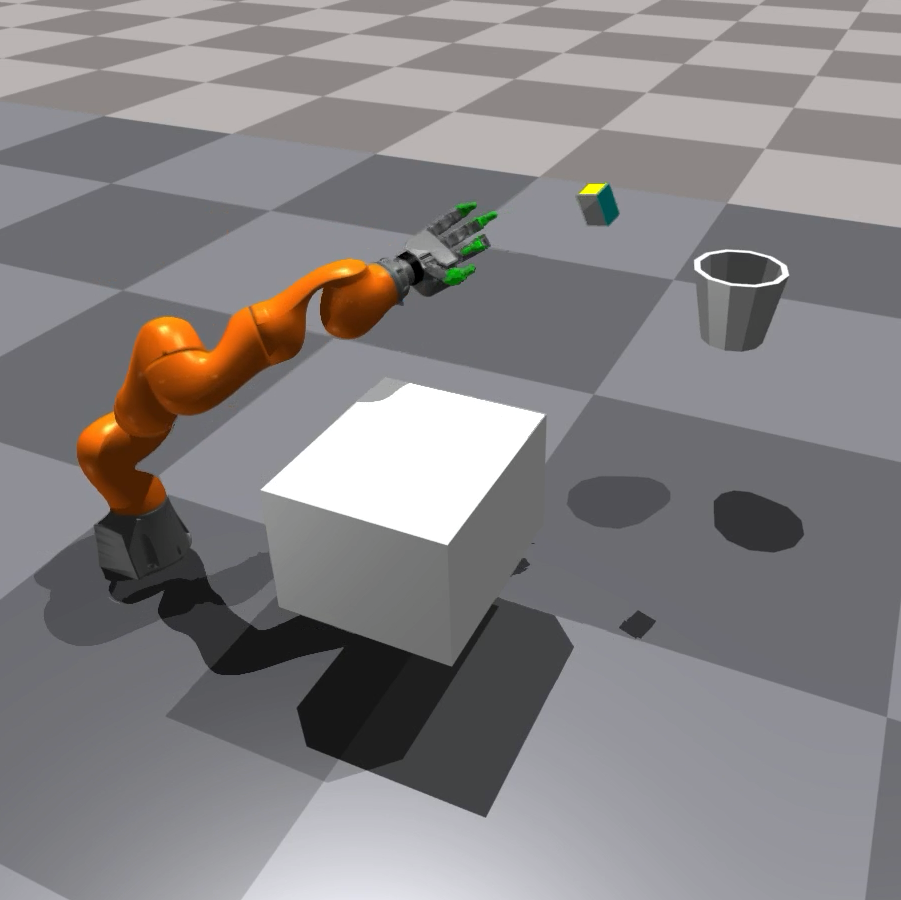}
    \includegraphics[width=0.245\linewidth]{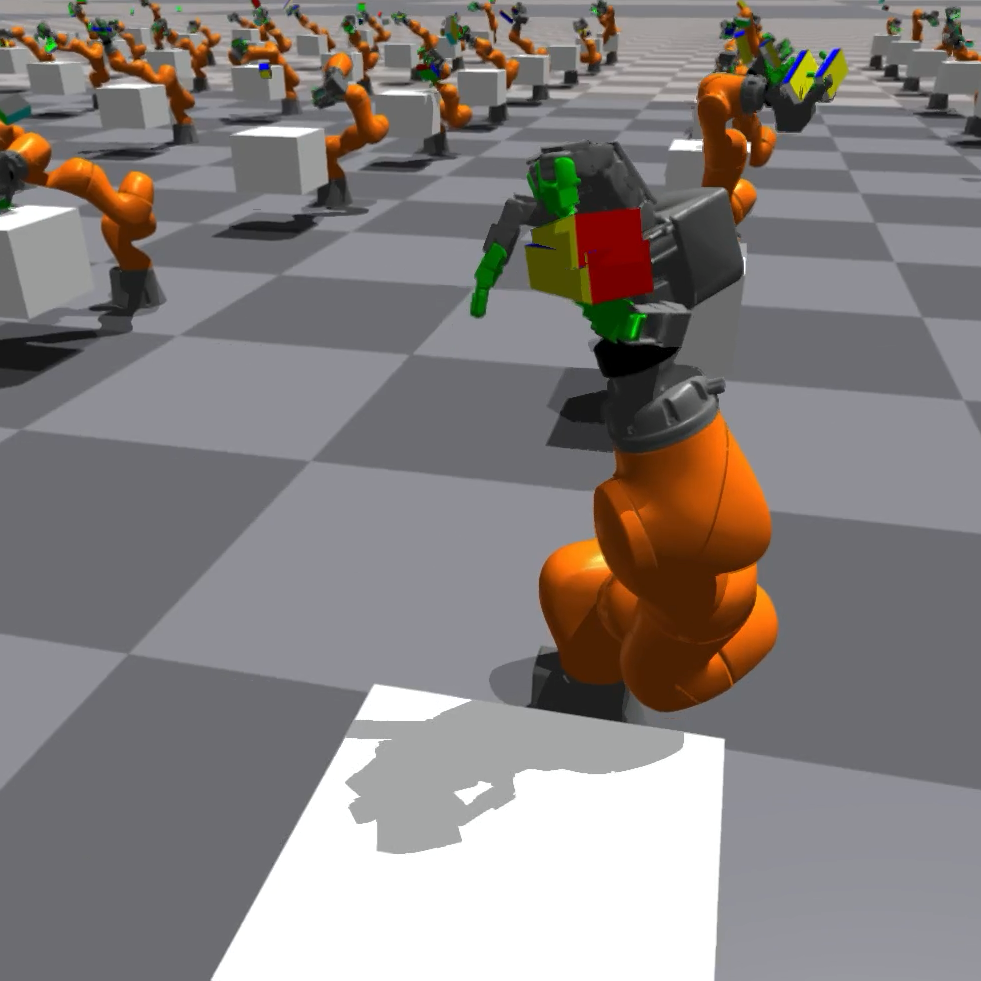}
    \includegraphics[width=0.245\linewidth]{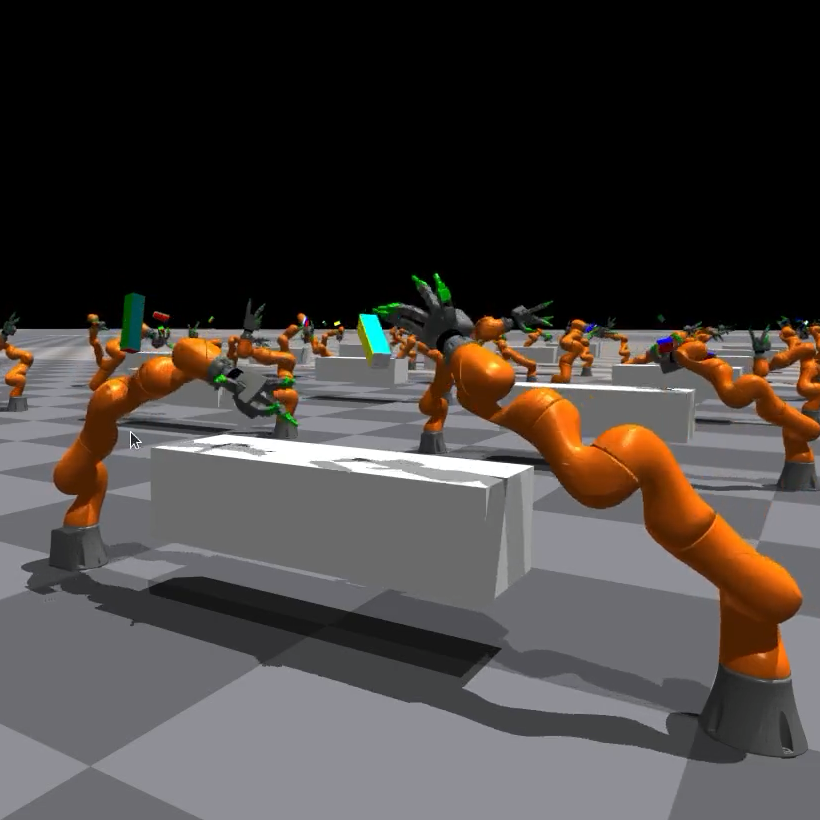}
    \captionof{figure}{Tasks trained with DexPBT. Left-to-right: \textit{regrasping}, \textit{throwing}, \textit{single-handed reorientation}, \textit{two-handed reorientation}.}
    \label{fig:teaser}
  \end{minipage}
    }
\makeatother

\maketitle

\begin{abstract}
In this work, we propose algorithms and methods that enable learning dexterous object manipulation using simulated one- or two-armed robots equipped with multi-fingered hand end-effectors. Using a parallel GPU-accelerated physics simulator (Isaac Gym), we implement challenging tasks for these robots, including regrasping, grasp-and-throw, and object reorientation. To solve these problems we introduce a decentralized Population-Based Training (PBT) algorithm that allows us to massively amplify the exploration capabilities of deep reinforcement learning. We find that this method significantly outperforms regular end-to-end learning and is able to discover robust control policies in challenging tasks. Video demonstrations of learned behaviors and the code can be found at the
\href{https://sites.google.com/view/dexpbt}{supplementary website}.
\end{abstract}

\IEEEpeerreviewmaketitle

\setcounter{figure}{1}

\section{Introduction}

\begin{figure*}[t]
\centering
    \includegraphics[width=0.95\linewidth]{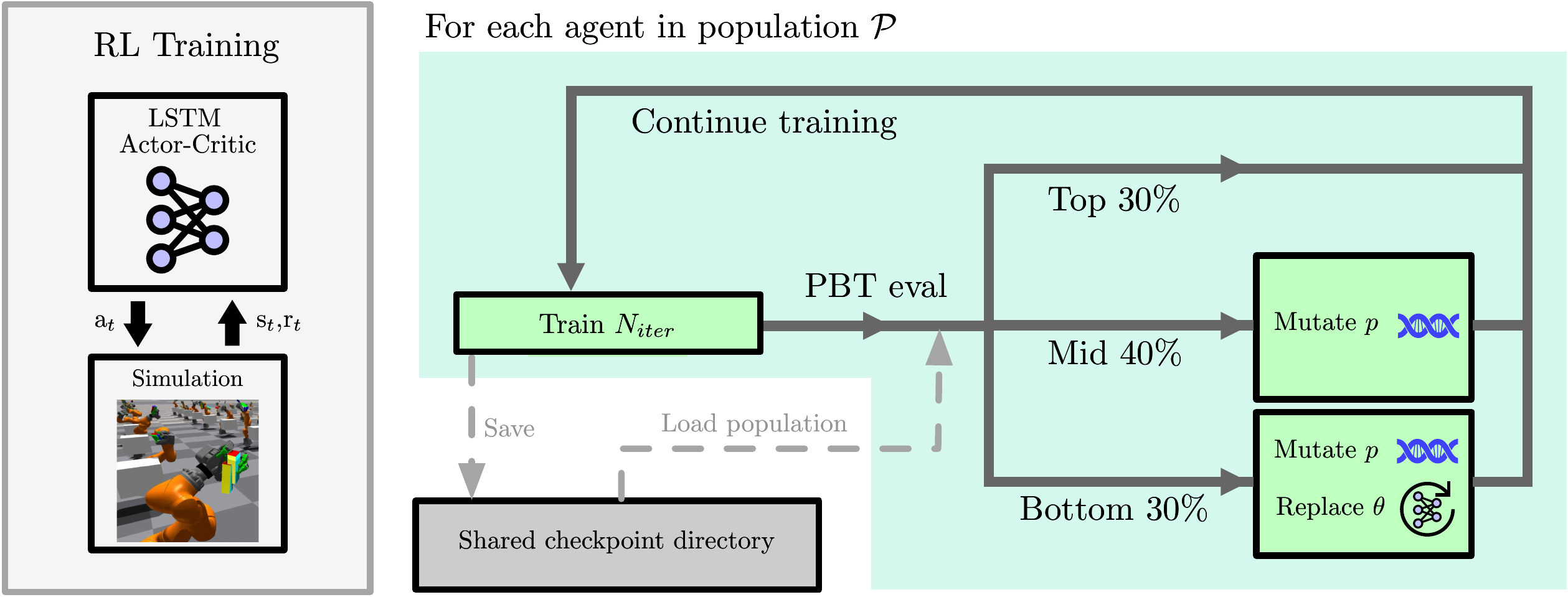}
    \caption{An illustration of the system used to solve our complex manipulation tasks using a combination of RL, highly parallelized robotic simulation, and Population Based Training (PBT).}
\label{fig:system-figure}
\end{figure*}

In recent years researchers have started to apply deep reinforcement learning methods in an increasing number of challenging continuous control domains. Some applications include impressive demonstrations of skill in virtual environments, such as playing football by directly controlling humanoid characters using joint torques~\cite{liu2022dmfootball}. In other domains such as agile drone flight~\cite{song_scaramuzza2021racing,molchanov2019simtomulti,batra2021swarm} and quadruped locomotion~\cite{joonholee2020quadruped,miki2022quadrupedperceptive}, control policies trained in simulation are deployed directly on the real hardware.

Learning-based approaches appear to be particularly promising in the domain of robotic manipulation. Traditional methods such as direct trajectory optimization, may struggle to model complex contact dynamics. Due to these difficulties, researchers working on manipulation problems typically focus their efforts on robotic arms with end-effectors that simplify contact handling, such as parallel jaw grippers~\cite{chebotar2019simopt,narang2022factory,matl2019ambidextrous}. Even though more capable human-like robotic hands have the potential to endow robots with far more advanced manipulation capabilities, this type of end-effector remains relatively unpopular due to the difficulty of controlling high degree-of-freedom~(DoF) systems in contact-rich environments.

With the advent of modern deep RL methods that use a large amount of data and computation, it has become possible to learn control policies for multi-fingered robotic hands. Systems like "Dactyl"~\cite{openai2020dactyl,openai2019rubiks} and "DeXtreme"~\cite{ahanda2022dextreme} have demonstrated how large-scale reinforcement learning in simulation can be used to obtain robust policies for complex in-hand object manipulation.

In this work, we extend this approach and apply it to a fully actuated hand-arm system: a four finger 16-DoF Allegro Hand mounted on a 7-DoF Kuka arm. We target a variety of manipulation tasks in simulated environment Isaac Gym~\cite{isaacgym}, such as regrasping, throwing, and reorientation. We then scale our learning method to train agents that control a pair of arms and hands with combined 46 degrees of freedom using a single neural network policy.

We observe that despite the relative success of straightforward end-to-end learning, our RL experiments are characterized by a high variance of results and dependency on initial conditions, especially in tasks that require exploration in the vast space of possible behaviors. To that end, we develop a Population-Based Training (PBT) algorithm~\cite{jaderberg2017pbt}, an outer optimization loop that can significantly amplify the exploration capabilities of end-to-end RL. In order to facilitate asynchronous learning in a volatile compute environment we implement a \textit{decentralized} version of the PBT algorithm that we can run without a central orchestrator instance and is robust to the disconnect of one or a few learners. We find that the PBT approach demonstrates improved performance in all scenarios over standard end-to-end learning and becomes the enabling factor for the successful training of ambidextrous agents that control two hand-arm systems simultaneously.

Our contributions can be summarized as follows:

\begin{itemize}
    \item We develop a framework that combines on-policy RL and decentralized Population Based Training with realistic GPU-accelerated robotic simulation and use this framework to train policies for dexterous object manipulation with high-DoF single and dual hand-arm systems.

    \item We introduce a staged reward function formulation and a PBT meta-objective to simplify and automate reward tuning and hyperparameter search.
    
    \item We release our environments, RL, and PBT code to facilitate further research in dexterous robotic manipulation (see the \href{https://sites.google.com/view/dexpbt}{supplementary website}).

\end{itemize}

\section{Related Work}

Producing control policies to perform complex, contact-rich tasks has been a long-standing challenge in robotics. Classical methods for this have focused on directly leveraging robot kinematics \citep{salisbury82, salisbury85, li89grasping}. While useful in free-space or pick-and-place style tasks, these methods struggle as the number of contacts grows.

Recently, various systems have leveraged learning-based methods to perform contact-rich robotic manipulation, achieving impressive results both in simulation and reality~\cite{levine16e2e, irpan18qtopt}. In particular, RL-based methods in simulation have shown the ability to learn complex and robust behaviors in realistic scenarios which transfer to the real world~\cite{joonholee2020quadruped, openai2020dactyl,ahanda2022dextreme}. The advent of high-throughput simulation on GPUs has improved the speed at which such tasks can be learned using RL~\citep{isaacgym, rudin22, allshire2021trifinger, raisim, freeman2021brax}.

Multiple prior projects have explored the problem of dexterous object manipulation. \citet{kumar2016localmodels} achieved in-hand object rotation with a Shadow Hand-like robotic hand using a model-based approach. \citet{openai2020dactyl} and \citet{ahanda2022dextreme} showed that it was possible to train the policy to do in-hand cube manipulation and even Rubik's cube \cite{openai2019rubiks} solving entirely in simulation and deploy the learned policy on the real robot. Other work has shown multiple tasks with free-floating hands~\cite{bimanual22} or with object grasping~\cite{Xiao2022}. \citet{matl2019ambidextrous} demonstrated methods for learning grasping policies with two robotic arms and different types of end-effectors such as suction cups and parallel jaw grippers. \citet{gupta2021resetfree} were able to learn object manipulation skills on a hand-arm system using off-policy reset-free reinforcement learning directly in the real world.
Our approach is most similar to \citet{ahanda2022dextreme}: we also take advantage of the massively parallel GPU-accelerated physics engine, and we add a Population-Based Training outer loop for improved exploration, automated reward function tuning, and hyperparameter optimization.

\citet{jaderberg2017pbt} popularized Population-Based Training in a variety of domains such as RL, adversarial learning, and machine translation, however PBT methods turned out to be particularly promising in deep RL where exploration is often the bottleneck. PBT provides a way to combine the exploration power of multiple learners and directs resources toward more promising behaviors. Since then, PBT algorithms have been exceptionally successful in RL for video game applications and helped to produce state-of-the-art agents for games such as Quake~\cite{jaderberg2018games}, Doom~\cite{petrenko2020sf}, and Starcraft II~\cite{vinyals19starcraft}.

Recently \citet{wan2022bayesian} augmented PBT-style methods with trust-region based Bayesian Optimization and were able to optimize both hyperparameters and model architectures simultaneously. \citet{flajolet2022fastbpt} proposed highly efficient JAX implementation of PBT that enables evaluation of multiple agents on one accelerator. Both \citet{wan2022bayesian} and \citet{flajolet2022fastbpt} demonstrated great results on standard continuous control benchmarks such as Half-Cheetah and Humanoid, in contrast, we apply our method in complex dexterous manipulation domains more similar to real robot settings. Additionally, our \textit{decentralized} PBT implementation (\mysec{sec:pbt}) makes it easy to use the algorithm in distributed compute environments such as Slurm clusters.

\section{Method}

\subsection{Problem Statement}

\begin{figure}[b]
\centering
\includegraphics[width=0.9\linewidth]{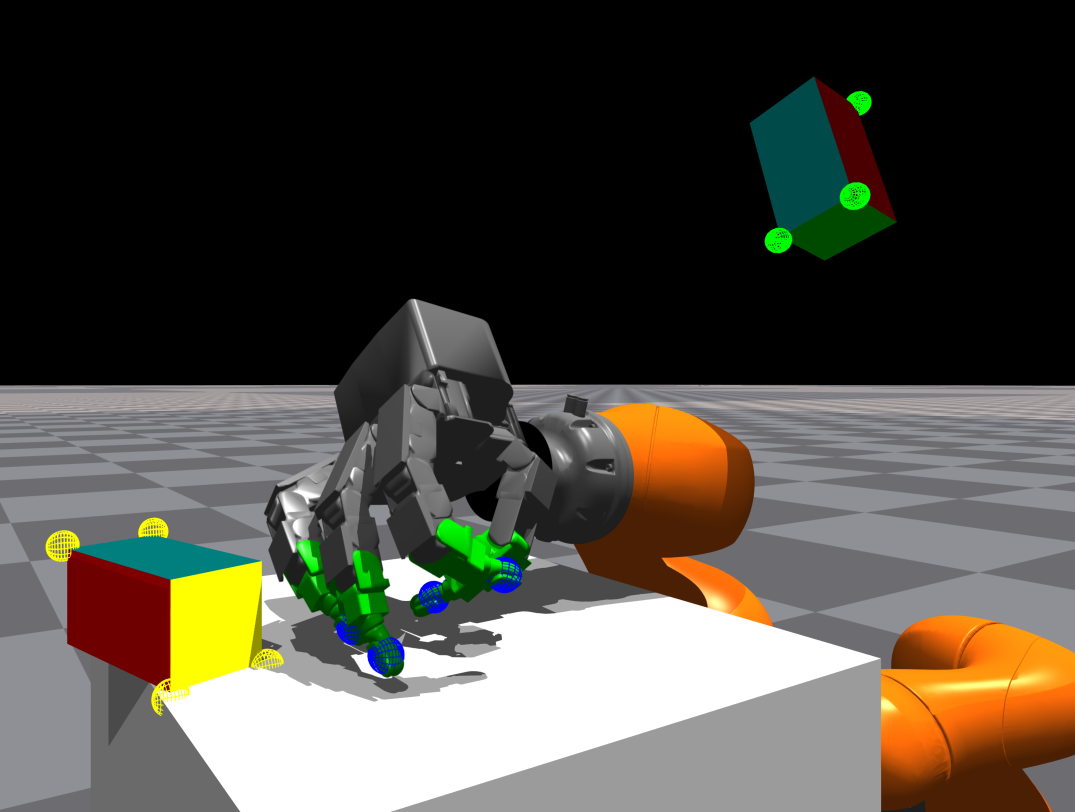}
\caption{
We use object keypoints (yellow) to represent both observed object position $x_{kp} \in \mathbb{R}^{3N_{kp}}$ and object shape. In reorientation tasks keypoints are also used to represent the desired object pose $x_{targ} \in \mathbb{R}^{3N_{kp}}$ (green). Observed fingertip locations are additionally rendered in blue.}
\label{fig:keypoints}
\end{figure}

A lot of problems in practical robotics require the robot to perform some notion of \textit{rearrangement}~\cite{batra2020rearrangement}, \ie bringing a given environment into a specified state. In household, factory, or warehouse environments many interesting tasks will involve a type of rearrangement that requires dexterous object handling and manipulation~\cite{okamura2000dexoverview}. In this work, we target a high-DoF anthropomorphic hand+arm system in the attempt to build towards the desired level of dexterity.

We focus on a problem that can be seen as a special case of rearrangement: \textit{single-object reposing}. This task requires changing the state of a single rigid body such that it matches the target position $x \in \mathbb{R}^3$ and, optionally, orientation $R \in SO(3)$. Dexterous single-object manipulation can be seen as an essential primitive required to perform general-purpose rearrangement. This task requires mastery of contact-rich grasping and in-hand manipulation and presents an exciting challenge for robotics research.

We approach dexterous manipulation as a discrete-time sequential decision-making process. At each step the controller observes environment state $s_t \in \mathbb{R}^{N_{obs}}$ (which includes the target object pose) and yields an action $a_t \in \mathbb{R}^{N_{dof}}$ specifying the desired angles of arm and finger joints. Whenever the object state matches the target within a specified tolerance, the attempt is considered successful, and the target state is reset. If the object is dropped during the attempt, or if the target state is not achieved within a time period $\tau$ we consider this attempt failed. The simulation proceeds until $N_{max}$ consecutive successes are reached or until the first failure. The performance on the task can thus be measured as a number of consecutive successes within the episode $N_{succ} \leq N_{max}$, as done in prior work~\cite{openai2020dactyl}.

Similar to \citet{allshire2021trifinger} we use $N_{kp}$ keypoints to represent the observed object pose $x_{kp} \in \mathbb{R}^{3N_{kp}}$. In tasks that require orientation matching, keypoints are also used to represent the desired object pose $x_{targ} \in \mathbb{R}^{3N_{kp}}$ (\Cref{fig:keypoints}). We consider such task successfully executed when all of $N_{kp}$ keypoints are within the tolerance threshold of their corresponding target locations ${\underset{i \in 1..N_{kp}}{\max} || x_{targ}^{(i)} - x_{kp}^{(i)} || \leq \varepsilon^*}$. For tasks that require only position matching we just require the object's center to be within $\varepsilon^*$ of the desired location.

The keypoint representation provides multiple benefits. It eliminates the need to tune tolerance thresholds and RL reward weights separately for rotation and translation. In addition to that, keypoints placed in a predefined pattern on the convex hull of the object allow the policy to observe the shape of the rigid body. In this work we use parallelepipeds with dimensions ranging from \SI{3}{\cm} to \SI{30}{\cm} as our manipulation targets, therefore a representation with $N_{kp}=4$ keypoints is sufficient to convey the shape information. 

\subsection{Scenarios}
\label{sec:scenarios}

We developed three variants of our object manipulation task that highlight different challenges in dexterous manipulation. These scenarios are \textit{regrasping}, \textit{grasp-and-throw}, and \textit{reorientation} (\Cref{fig:teaser}). At the beginning of each episode, the object appears in a random position on the table and the hand-arm system is reset to a random state $s_{robot} \in \mathbb{R}^{2 N_{dof}}$ consisting of random joint angular velocities and initial angles within the DoF limits.
The episode then proceeds according to one of the following scenarios.

The \textit{regrasping} task demands that the agent grasp the object, pick it up from the table, and hold it in a specified location for a duration of time, after which both object and target positions are reset. To succeed in this scenario the control policy must develop stable grasps that minimize the probability of dropping the object during the attempt.

In \textit{grasp-and-throw} the task is to pick up the object and displace it into a container that can be outside of manipulator's reach. This scenario requires aiming for the container and throwing the object a significant distance. Here we test the ability of the control policy to understand the dynamic aspects of object manipulation: successful execution of this task requires releasing the grip at the right point of the trajectory, giving the object just the right amount of momentum to direct it towards the goal. After each attempt we reset the positions of the object and the container to make sure that the policy is able to complete the task repeatedly for a diverse set of initial conditions.

The \textit{reorientation} task provides perhaps the most difficult object manipulation challenge among the chosen scenarios. The goal is to grasp the object and consecutively move it to different target positions and orientations. This scenario requires maintaining a stable grip for minutes of simulated time, fine control of the joints of the robotic arm, and occasional in-hand rotation when the reorientation cannot be performed by merely using the affordances of the Kuka arm (\href{https://www.youtube.com/embed/wM2WuKUtVp8}{video demonstration}). Thus formulated, the reorientation task includes elements seen in previous work such as dexterous in-hand manipulation \cite{openai2019rubiks,ahanda2022dextreme}, but extends the capability of the manipulator to perform reposing in much larger volume.
While regrasping and grasp-and-throw tasks only require matching the target position, reorientation scenario additionally demands that object rotation matches the target.

In both regrasping and reorientation the final required tolerance is $\varepsilon^* = \SI{1}{\cm}$, thus demanding very precise control. For grasp-and-throw we use $\varepsilon^* = \SI{7.5}{\cm}$ since the main focus is on landing the object into the container, not the precise positioning within it.

\mypara{Dual-Arm Scenarios}.
In the attempt to find the limits of end-to-end learning for continuous control, we introduce versions of regrasping and reorientation scenarios for two hand-arm systems. It is likely not a coincidence that humans, sculpted into a form optimised for object manipulation by evolution, wield not one but a pair of arms and hands. Thus solving object reposing with two high-DoF manipulators in simulation can be seen as a milestone on the path towards future robotic systems that can match or exceed human dexterity.

To produce dual-arm scenarios we double the number of simulated robots per task for the total of 46 degrees of freedom (\Cref{fig:teaser}, on the right). We change the sampling of initial and target object positions in a way that guarantees that the task cannot be solved by any one robot. Thus, a complete solution requires grasping the object, passing the object from one hand to another, as well as in-hand manipulation, combining the most challenging elements from all scenarios. We extend both observation (\mytable{table:policy-inputs}) and action space ($a_t \in \mathbb{R}^{N_{dof} * N_{arm}}, N_{arm}=2$) to allow a single policy to control both manipulators.

\begin{table}
    \footnotesize
    \centering
    \caption{Actor/critic observations and their dimensionality. Here $N_{arm} = 1$ for single-arm and $N_{arm} = 2$ for dual-arm tasks, and $N_{kp} = 4$.}
    \begin{sc}
    \begin{tabular}{@{}lr@{}}
        \toprule
        \rowcolor[HTML]{D4F7EE}
        \textbf{Input} & \textbf{Dimensionality} \\
        \midrule
        Joint angles & $23D * N_{arm}$ \\
        Joint velocities & $23D * N_{arm}$ \\
        Hand position & $3D * N_{arm}$ \\
        Hand rotation & $3D * N_{arm}$ \\
        Hand velocity & $3D * N_{arm}$ \\
        Hand angular velocity & $3D * N_{arm}$ \\
        Fingertip positions & $12D * N_{arm}$ \\
        \midrule
        Object keypoints rel. to hand & $3D * N_{kp} * N_{arm}$ \\
        Object keypoints rel. to goal & $3D * N_{kp}$ \\
        Object rotation & $4D$ (quaternion) \\
        Object velocity & $3D$ \\
        Object angular velocity & $3D$ \\
        Object dimensions & $3D$ \\
        \midrule
        $\1_{picked}$ & $1D$ \\ 
        $d_{closest}$ & $1D$ \\ 
        $\hat{d}_{closest}$ & $1D$ \\ 
        \midrule \midrule
        Total, one arm tasks & $110D$ \\
        Total, dual-arm tasks & $192D$ \\

        \bottomrule
    \end{tabular}
    \end{sc}
\label{table:policy-inputs}
\end{table}

\subsection{Reinforcement Learning}
\label{sec:rl}

We formalize the problem as a Markov Decision Process (MDP) where the agent interacts with the environment to maximize the expected episodic discounted sum of rewards $\mathop{\mathbb{E}}[\sum_{t=0}^{T} \gamma^{t}r(s_{t},a_{t})]$. We use Proximal Policy Optimization algorithm~\cite{schulman2017ppo} to simultaneously learn the policy $\pi_{\theta}$ and the value function $V_{\theta}^{\pi}(s)$, both parameterized by a single parameter vector $\theta$. The model architecture is an LSTM \cite{hochreiter1997lstm} followed by a 3-layer MLP. Even though agents can observe all relevant parts of the environment state, we decided to follow prior work~\cite{openai2020dactyl,openai2019rubiks} and train recurrent models because any future real-world deployment will necessarily involve partial observability and require memory for test-time system identification.

The policy is trained using experience simulated in Isaac Gym~\cite{isaacgym}, a highly parallelized GPU-accelerated physics engine. To process high volume of data generated by this simulator we use an efficient PPO implementation~\cite{rl-games} which keeps the computation graph entirely on the GPU. Combined with the minibatch size of $2^{15}$ transitions, this allows us to maximize the hardware utilization and learning throughput.

We utilize normalization of observations, advantages, and TD-returns~\cite{schulman2016gae} to make the algorithm invariant to absolute scale of observations and rewards. We also use an adaptive learning rate algorithm that maintains a constant KL-divergence $D_{KL}(\pi|\pi_{old})$ between the current policy $\pi_{\theta}$ and the behavior policy $\pi_{\theta_{old}}$ that collected the rollouts. The learning rate is reduced by a factor of $1.5$ whenever $D_{KL}$ exceeds the threshold by the end of the training iteration, and is multiplied by $1.5$ when $D_{KL}$ falls below the threshold.

Both regrasping and reorientation demand very precise control ($\varepsilon^*=\SI{1}{\cm}$). Because of that, agents almost never encounter successul task execution early in the training. In order to create a smooth learning curriculum we adaptively anneal the tolerance from a larger initial value $\varepsilon_{0}=\SI{7.5}{\cm}$. We periodically check if the policy crossed the performance threshold $N_{succ} > 3$, and in this case we decrease the current success tolerance until it reaches the final value $\varepsilon^*$: $\varepsilon \gets \max(0.9 \varepsilon, \varepsilon^*)$.

\mypara{Reward function.} For the successful application of any reinforcement learning method, the reward should be dense enough to facilitate exploration yet should not distract the agent from the sparse final objective
(which in our case is to maximize the number of consecutive successful manipulations).

We propose a reward function that naturally guides the agent through a sequence of motions required to complete the task, from reaching for the object to picking it up and moving it to the final location:
\begin{equation}
\label{eq:reward_function}
r(s,a) = r_{reach}(s) + r_{pick}(s) + r_{targ}(s) - r_{vel}(a).
\end{equation}

Here $r_{reach}$ rewards the agent for moving the hand closer to the object at the start of the attempt:
\begin{equation}
\label{eq:reach_rew}
    r_{reach} = \alpha_{reach} * \max(d_{closest} - d, 0),
\end{equation}
where both $d$ and $d_{closest}$ are distances between the end-effector
and the object, $d$ is the current distance, and $d_{closest}$ is the closest distance achieved during the attempt so far. In dual-arm scenarios we calculate distance $d$ for the end-effector that's closer to the object.

Component $r_{pick}$ rewards the agent for picking up the object and lifting it off the table:
\begin{equation}
\label{eq:pick_rew}
    r_{pick} = (1 - \1_{picked}) * \alpha_{pick} * h_t + r_{picked}.
\end{equation}
In this equation $\1_{picked}$ is an indicator function which becomes 1 once the height of the object relative to the table $h_t$ exceeds a predefined threshold of \SI{15}{\cm}. At this moment the agent receives an additional sparse reward $r_{picked}$.
Once the object is picked up, $r_{targ}$ rewards the agent for moving the object closer to the target state:
\begin{equation}
\label{eq:target_rew}
    r_{targ} = \1_{picked} * \alpha_{targ} * \max(\hat{d}_{closest} - \hat{d}, 0) + r_{success}.
\end{equation}

In the reorientation task $\hat{d} = \underset{i \in 1..N_{kp}}{\max} || x_{targ}^{(i)} - x_{kp}^{(i)} ||$ is the maximum distance between corresponding pairs of object and target keypoints, while in tasks that do not require orientation matching $\hat{d}$ is simply the distance between the object center and the target location; in both cases $\hat{d}_{closest}$ is the smallest $\hat{d}$ achieved during the attempt so far. A large sparse reward $r_{success}$ is added when the desired position and/or orientation is reached, $\hat{d} = \hat{d}_{closest} \leq \varepsilon^*$. 

Finally, $r_{vel}$ in Eq.~(\ref{eq:reward_function}) is a simple joint velocity penalty that can be tuned to promote smoother movement, and in \Cref{eq:reach_rew,eq:pick_rew,eq:target_rew} $\alpha_{reach},\alpha_{pick},\alpha_{targ}$ are relative reward weights. Note that  we apply virtually the same reward function in all scenarios, sans the minor differences in $\hat{d}$ calculation.

Overall, our reward formulation follows a sequential pattern: the reward components $r_{reach}$, $r_{pick}$ and $r_{targ}$ are mutually exclusive and do not interfere with each other. For example: by the time the hand approaches the object, the component $r_{reach}$ is exhausted since $d = d_{closest} = 0$, therefore $r_{reach}$ does not contribute to the reward for the remainder of the trajectory. Likewise, $r_{pick} \neq 0$ if and only if $r_{targ} = 0$ and vice versa due to the indicator function $\1_{picked}$. The fact that only one major reward component guides the motion at each stage of the trajectory makes it easier to tune the rewards and avoid interference between reward components. This allows us to avoid many possible local minima: for example, if $r_{pick}$ and $r_{targ}$ are applied together, depending on the relative reward magnitudes the agent might choose to slide the object to the edge of the table closer to the target location to maximize $r_{targ}$ and cease further attempts to pick it up and solve the problem for the fear of dropping the object.

In addition to that, rewards in \Cref{eq:reach_rew,eq:target_rew} have a predefined maximum total value depending on the initial distance between the hand and the object, and the object and the target respectively. This eliminates an entire class of reward hacking behaviors where the agent would remain close but not quite at the goal to keep collecting the proximity reward. In our formulation, only movement towards the goal is rewarded while mere proximity to the goal is not.

\begin{table}[t]
    \footnotesize
    \centering
    \caption{RL hyperparameters and reward function coefficients. Rightmost column provides an example of the highest scoring agent's final parameter values for a single dual-arm reorientation PBT experiment (parameters not optimized by PBT are omitted). }
    \begin{sc}
    \resizebox{\linewidth}{!}{%
    \begin{tabular}{@{}lcc@{}}
        \toprule
        \rowcolor[HTML]{D4F7EE}
        \textbf{Parameter} & \textbf{Initial value} & \makecell{\textbf{PBT-optimized} \\ \textbf{value}} \\
        \midrule
        LSTM size & 768 & - \\
        MLP layers & [768,512,256] & - \\
        Nonlinearity & ELU & - \\
        \midrule
        Discount factor $\gamma$ & 0.99 & 0.9888 \\
        GAE discount $\lambda$~\cite{schulman2016gae} & 0.95 & - \\
        Learning rate & Adaptive (Sec \ref{sec:rl}) & - \\
        Adapt. LR $D_{KL}(\pi|\pi_{old})$ & 0.016 & 0.01432 \\
        Gradient norm & 1.0 & 1.028 \\
        PPO-clip $\epsilon$ & 0.1 & 0.2564 \\
        Critic loss coeff. & 4.0 & 5.188 \\
        Entropy coeff. & 0 & - \\
        Num. agents & 8192 & - \\
        Minibatch size & 32768 & - \\
        Rollout length & 16 & - \\
        Num. PPO epochs & 2 & 1 \\
        \midrule
        \normalsize{$\alpha_{reach}$} & 50 & 74.8 \\
        \normalsize{$\alpha_{pick}$} & 20 & 22.1 \\
        \normalsize{$r_{picked}$} & 300 & 414.4 \\
        \normalsize{$\alpha_{targ}$} & 200 & 263.5 \\
        \normalsize{$r_{success}$} & 1000 & 1322.7 \\
        \bottomrule
    \end{tabular}
    }
    \end{sc}
\label{table:parameters}
\end{table}

\mypara{Observations and actions.} In our experiments both actor $\pi_{\theta}$ and critic $V_{\theta}^{\pi}(s)$ observe environment state directly, including joint angles and velocities, positions of fingertips, object rotation, velocity, and angular velocity. Additionally, keypoint positions and object dimensions provide information about the object shape. \Cref{table:policy-inputs} lists all observations available to the agent and their corresponding dimensionalities.

The policy $\pi_{\theta}$ outputs two vectors $\mu,\sigma \in \mathbb{R}^{N_{dof} * N_{arm}}$ which are used as parameters of $N_{dof} * N_{arm}$ independent Gaussian probability distributions. Actions are sampled from these distributions $a \sim \mathcal{N}(\mu, \sigma)$, clipped to corresponding joint limits and interpreted as target joint angles. Then a PD controller yields joint torques in order to get joints to the target angles specified by the policy.

It is unrealistic to expect that all of the mentioned observations are available on the real robot. In this case, we propose using all aforementioned observations for the critic during training, while the policy only receives a subset of observations that can be obtained from \ie a vision system, also known as asymmetric actor-critic approach (see~\cite{ahanda2022dextreme,openai2019rubiks}).

\subsection{Population-Based Training}
\label{sec:pbt}

A contact-rich continuous control problem with up to 192 observation dimensions (\Cref{table:policy-inputs}) and up to 46 action dimensions can be exceptionally challenging even for modern RL algorithms~\cite{rl-games}. The main challenge is exploration: from the large number of possible behaviors that maximize rewards early in the training only relatively few lead to high-performance solutions at convergence.

Another dimension of complexity when using learning methods is hyperparameter tuning. Any reward shaping scheme contains a number of coefficients that need to be carefully balanced in order to maximize the objective. In addition to that, modern RL algorithms have a substantial number of settings, such as learning rate, number of training epochs on each set of collected trajectories, relative magnitudes of actor and critic losses, and so on. Choosing these parameters can be quite challenging and heavily relies on the expertise of engineers and researchers.

In order to mitigate these problems we employ a Population-Based Training approach~\cite{jaderberg2017pbt}. The core idea is akin to an evolutionary algorithm: we train a population of agents $\mathcal{P}$, perform mutation to generate promising hyperparameter combinations, and use selection to prioritize agents with superior performance. Each agent $(\theta_i, p_i) \in \mathcal{P}$ is characterized by a parameter vector $\theta_i$ and a set of hyperparameters $p_i$, which includes settings of the RL algorithm as well as reward coefficients $\alpha_{reach}, \alpha_{pick}, \alpha_{targ}, r_{picked}, r_{success}$ (see \mytable{table:parameters}). Periodically (after training on $N_{iter}$ environment transitions), each agent is evaluated to obtain the target performance metric $r_{meta}$ and the population is sorted according to this metric into three subsets $\mathcal{P}_{top}, \mathcal{P}_{mid}, \mathcal{P}_{bottom} \subset \mathcal{P}$ with cardinality equal to 30\%, 40\%, and 30\% of $\mathcal{P}$ respectively (see \Cref{fig:system-figure} and \Cref{alg:pbt}). Note that for performance reasons we use $r_{meta}$ measured at the end of the training iteration to approximate a separate evaluation procedure $\textsc{eval}(\theta)$.

Agents that belong to the highest-performing subset $\mathcal{P}_{top} \subset \mathcal{P}$ continue training without interruption. Agents in the middle 40\% of the population $\mathcal{P}_{mid}$ undergo hyperapameter mutation and continue training. Underperforming agents $(\theta, p) \in \mathcal{P}_{bottom}$ are discarded and get replaced with a randomly sampled high-performing agent $(\theta^*, p^*) \sim \mathcal{P}_{top}$ with mutated copy of hyperparameters $p^*$. This algorithm directs computational resources towards more promising agents in $\mathcal{P}_{top}$ and explores numerous hyperparameter combinations, maximizing exploration (as 70\% of the population constantly undergoes mutation).

The hyperparameter mutation scheme is kept simple: at each iteration of mutation each regular floating-point hyperparameter has a $\beta_{mut}$ probability to be multiplied or divided by random number sampled from the uniform distribution $\mu \sim \mathcal{U}(\mu_{min}, \mu_{max})$ (see \Cref{alg:mutation} and \Cref{table:pbt_parameters} for details). Note that discrete hyperparameters (such as the number of PPO epochs) and parameters with limited scope (such as discount factor $0 < \gamma < 1$) require slightly different mutation rules.

Some hyperparameter mutations may temporarily lead to decreased performance, \eg even relatively small discount factor mutations lead to significant changes in the distribution of TD-returns, thus decreasing the accuracy of the critic. Yet despite this temporary disadvantage these changes may lead to long-term benefits. To give agents a chance to adapt to altered parameters we temporarily pause PBT updates for this agent for $N_{adapt}=5 \times 10^7$ steps. Similar to \citet{jaderberg2018games} and \citet{petrenko2020sf} we also enable PBT only $N_{start}=2 \times 10^8$ steps after the beginning of training in order to promote population diversity. This helps us prevent the situation where the entire population is filled by copies of one particularly lucky seed.

\begin{algorithm}[t]
\small
\caption{Population-Based Training}
\label{alg:pbt}
\begin{algorithmic}[1]
\Require $\mathcal{P}$ (initial population, $\theta, p$ sampled randomly)
\For{$(\theta, p) \in \mathcal{P}$} (async. and decentralized)
\While{not end of training}
\State $\theta \gets \Call{train}{\theta, p}$ \Comment{Do RL for $N_{iter}$ steps}
\State $r_{meta} \gets \Call{eval}{\theta}$
\State $\mathcal{P}_{top}, \mathcal{P}_{mid}, \mathcal{P}_{bottom} \subset \mathcal{P}$ \Comment{Sort $\mathcal{P}$ according to $r_{meta}$}
\State $(\theta^*, p^*) \sim \mathcal{P}_{top}$ \Comment{Get agent from top 30\%}
\If{$(\theta, p) \in \mathcal{P}_{bottom}$}
    \State $p \gets \Call{mutate}{p^*}$
    \State $\theta \gets \theta^*$ \Comment{Replace weights}
\ElsIf{$(\theta, p) \in \mathcal{P}_{mid}$}
    \State $p \gets \Call{mutate}{p}$
\EndIf
\EndWhile
\EndFor
\State \textbf{return} $\theta_{best} \in \mathcal{P}$ \Comment{Agent with the highest $r_{meta}$}
\end{algorithmic}
 
\end{algorithm}

\begin{table}[t]
    \footnotesize
    \centering
    \caption{Parameters of the PBT algorithm. }
    \begin{sc}
    \begin{tabular}{@{}lcc@{}}
        \toprule
        \rowcolor[HTML]{D4F7EE}
        \textbf{Parameter} & \textbf{Value} \\
        \midrule
        Population size $|\mathcal{P}|$ & 8, 16, or 32 agents \\
        Population split $|\mathcal{P}_{top}|, |\mathcal{P}_{mid}|, |\mathcal{P}_{bottom}|$ & 30\%, 40\%, 30\% of $|\mathcal{P}|$ \\
        Initial delay $N_{start}$ & 200,000,000 env. steps \\
        Burn-in after mutation $N_{adapt}$ & 50,000,000 env. steps \\
        Normal PBT periodicity $N_{iter}$ & 20,000,000 env. steps \\
        Mutation probability $\beta_{mut}$ & 0.2 \\
        Min. perturbation $\mu_{min}$ & 1.1 \\
        Max. perturbation $\mu_{min}$ & 1.5 \\
        
        \bottomrule
    \end{tabular}
    \end{sc}
\label{table:pbt_parameters}
\end{table}

\mypara{Meta-objective}. One advantage of PBT is that it introduces an outer optimization loop that can meta-optimize for a final sparse scalar objective as opposed to inner RL loop which balances various dense reward components. Although meta-optimizing for $N_{succ}$ is an obvious choice, the adaptive tolerance annealing described in \mysec{sec:rl} creates a complication: different agents in the population can have different current values of success tolerance $\varepsilon$ and therefore cannot be compared directly as it is much easier to achieve high $N_{succ}$ at looser tolerance.

To address this issue, we define our meta-optimization objective that takes both $N_{succ}$ and $\varepsilon$ into account:
\begin{equation}
\label{eq:meta_objective}
\begin{cases}
    r_{meta} = \frac{\varepsilon_{0} - \varepsilon}{\varepsilon_{0} - \varepsilon^*} + 0.01 * N_{succ} & \text{if $\varepsilon > \varepsilon^*$} \\
    r_{meta} = 1 + N_{succ} & \text{if $\varepsilon = \varepsilon^*$} \\
\end{cases}
\end{equation}
Until the target tolerance $\varepsilon^*$ is reached, this objective is dominated by the term $0 \leq \frac{\varepsilon_{0} - \varepsilon}{\varepsilon_{0} - \varepsilon^*} \leq 1$ which is maximized when $\varepsilon$ approaches $\varepsilon^*$. After the desired tolerance is reached the maximization of $N_{succ}$ is prioritized.

\begin{algorithm}[b]
\small
\caption{Hyperparameter mutation subroutine.}
\label{alg:mutation}
\begin{algorithmic}[1]
\Function{mutate}{$p$} \Comment{Hyperparameter vector $p$}
\For{$j \gets 1$ to $len(p)$}
\State $z' \sim \mathcal{U}(0,1)$
\If{$z_{mut} < \beta_{mut}$} \Comment{Mutation probability}
\State $\mu \sim \mathcal{U}(\mu_{min}, \mu_{max})$ \Comment{Mutation amount}
\State $z_{dir} \sim \mathcal{U}(0,1)$ \Comment{Mutation direction}
\If{$z_{dir} < 0.5$}
\State $p_j \gets p_j * \mu$
\Else
\State $p_j \gets p_j / \mu$
\EndIf
\EndIf
\EndFor
\State \textbf{return} $p$
\EndFunction
\end{algorithmic}
 
\end{algorithm}

\begin{figure*}[t!]
\centering
\begin{subfigure}{.75\textwidth}
  \centering
  \includegraphics[width=\textwidth]{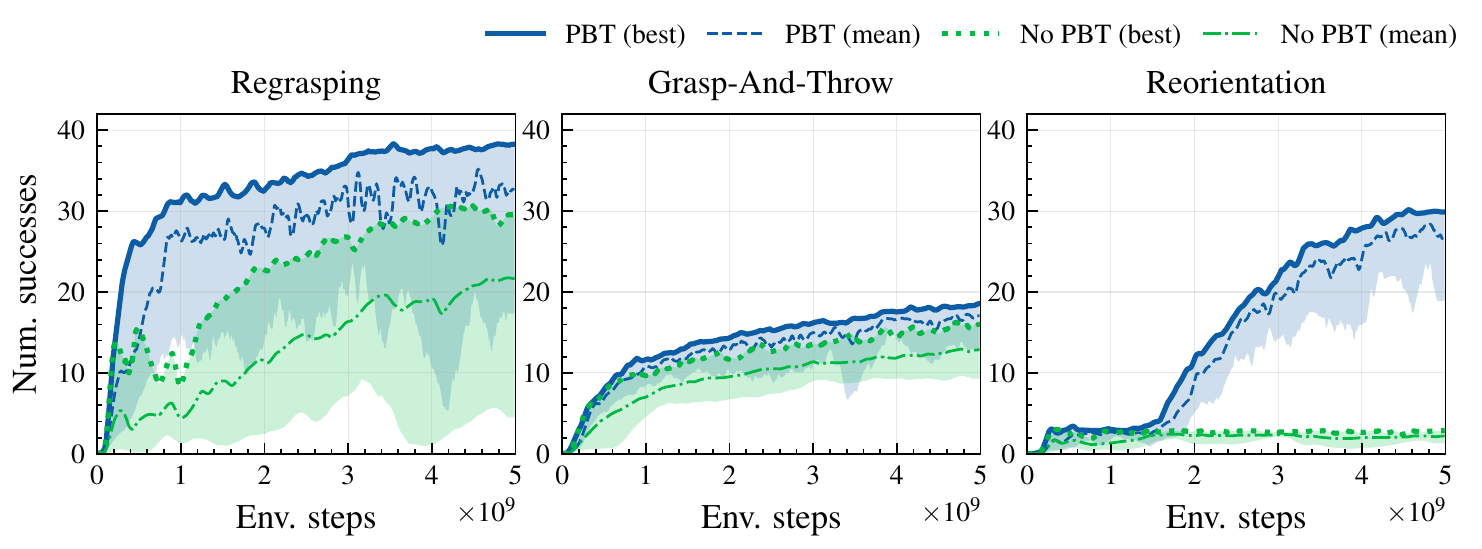}
\end{subfigure}%
\begin{subfigure}{.25\textwidth}
  \centering
  \includegraphics[width=\textwidth]{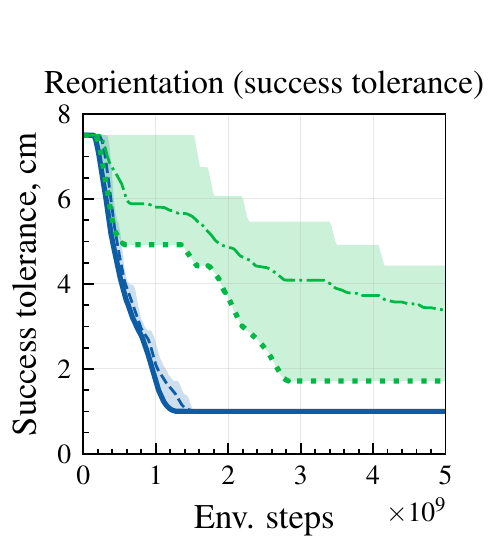}
\end{subfigure}
\caption{Training curves with and without PBT for \textbf{single-arm + hand tasks}. Shaded area is between the best and the worst policy among 8 agents in $\mathcal{P}$ or 8 seeds in non-PBT experiments.  }
\label{fig:single-arm-pbt}
\end{figure*}

\mypara{Decentralized PBT}. A trait that characterizes our implementation is a complete lack of any central orchestrator, typically found in other algorithms~\cite{petrenko2020sf,liaw2018tune}. Instead, we propose a completely \textit{decentralized} PBT architecture (see ~\Cref{fig:system-figure}). In such architecture, each agent is responsible for execution of its own part of \Cref{alg:pbt}, including finding its ranking in $\mathcal{P}$, mutating its own hyperparameters $p$, or replacing self with a copy of another agent.

In our implementation, agents in $\mathcal{P}$ interact exclusively through low-bandwidth access to a shared network directory containing histories of agent checkpoints and performance metrics for each agent. This eliminates the need for any other form of network communication or message passing between instances and makes it exceptionally easy to deploy the system, as most popular clusters provide some kind of shared filesystem. Our approach allows us to run exactly the same code with NFSv4 on Slurm, sshfs on NGC, S3 on AWS, or local filesystem on a multi-GPU server, provided that a shared workspace folder is mounted and accessible via a predefined path.

The lack of any central controller not only removes a point of failure, but also allows training in a volatile compute environment such as a contested cluster where some jobs can remain in queue for a long time. In this case, agents that start training later will be at a disadvantage when compared to other members of $\mathcal{P}$ that started earlier. To allow such agents to contribute to the population, we compare their performance only to historical checkpoints (of more advanced agents) that correspond to the same amount of collected experience.

It is almost inevitable that some learners may be lost during training, \eg due to random hardware issues. We require no special treatment for such agents and simply use the evaluation result from the latest available checkpoint. Although reduced number of active agents will almost certainly hinder the exploration capabilities of the algorithm, empirically we observe that decentralized PBT is quite robust to loss of one or a few agents, especially with larger population sizes. Alternatively, is is also possible to detect agent disconnects in a decentralized way (\eg by the lack of recent updates) and exclude such agents from the selection process, effectively reducing $|\mathcal{P}|$.

\section{Experiments}
\label{sec:experiments}

We conduct our experiments using instances with 8 CPU cores and a single Nvidia V100 GPU with 16 Gb of VRAM. Using the Isaac Gym engine~\cite{isaacgym} we are able to simulate 8192 parallel environments on each GPU. Combined with a GPU-based vectorized RL implementation rl\_games ~\cite{rl-games}, this allows us to reach training throughput of $5 \times 10^4$ samples per second for single-arm tasks and $3.5 \times 10^4$ samples per second for dual-arm scenarios. At this rate, to train successful policies on $5 \times 10^9$ environment transitions, it takes 30 hours for single- and 40 hours for dual-arm tasks (PBT and non-PBT training having almost equivalent per-node throughput).

For non-PBT experiments we simply train policies starting from multiple random seeds, each trained on a single instance. In our PBT experiments we use $|\mathcal{P}|$ separate 1-GPU instances that exchange information using low-bandwidth access to a shared directory (in our case, an NFS/sshfs shared folder on a Slurm/NGC cluster). \Cref{table:parameters} lists RL parameters and reward shaping coefficients used in our experiments.

\begin{figure}[b]
\centering
\includegraphics[width=1.0\linewidth]{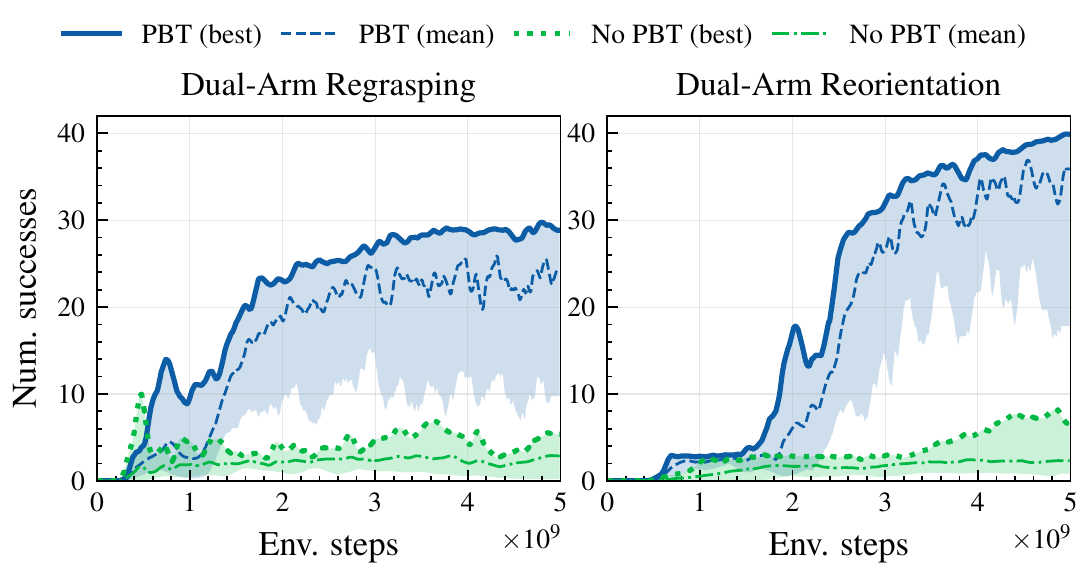}
\caption{Training curves with and without PBT for \textbf{dual-arm + hand tasks}. Shaded area is between the best and the worst policy among 8 agents in $\mathcal{P}$ or 8 seeds in non-PBT experiments.}
\label{fig:dual-arm-pbt}
\end{figure}

\begin{figure*}
\centering
\includegraphics[width=1.0\linewidth]{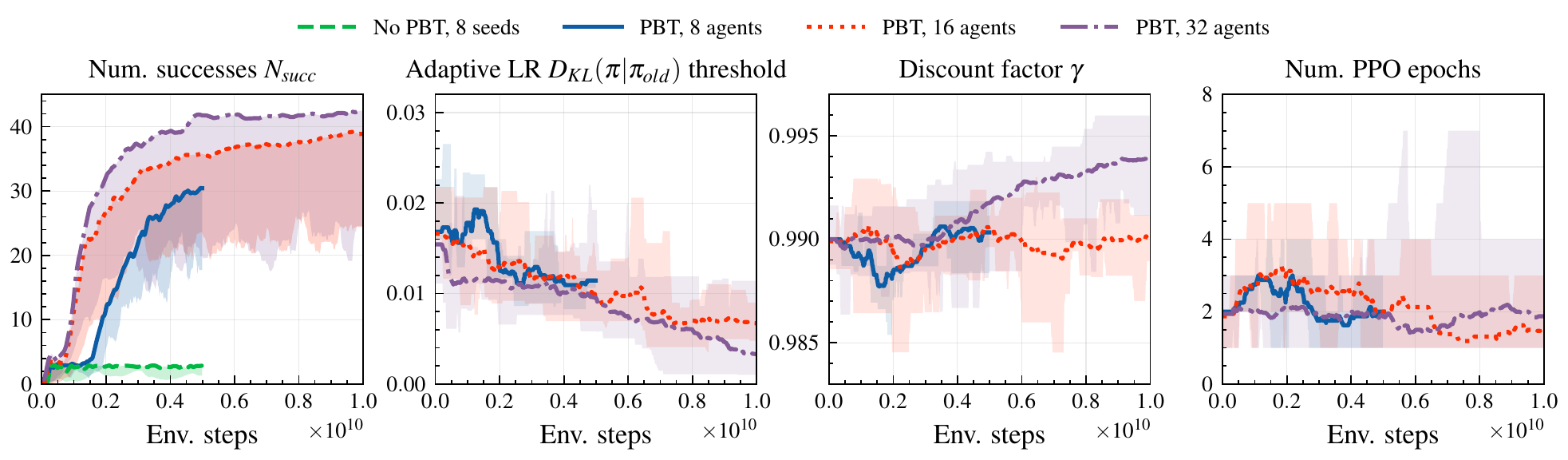}
\caption{Extended PBT experiments on Single-Arm Reorientation task with $|\mathcal{P}|=16$ and $|\mathcal{P}|=32$ agents, trained on 10 billion environment steps. 8-agent PBT run and non-PBT PPO (see \Cref{fig:single-arm-pbt}) trained to 5 billion steps superimposed for comparison. Leftmost plot shows the performance of the best agent in $\mathcal{P}$ and the shaded area is between the best and the worst policy according to $N_{succ}$. Additional plots show hyperparameter schedules discovered by PBT: adaptive learning rate KL-divergence threshold (\Cref{sec:rl}), RL discount factor $\gamma$, and the number of PPO epochs per training iteration (see \Cref{table:parameters}). We highlight the mean population value and shade the area between the minimum and the maximum value of the hyperparameter in $\mathcal{P}$. }
\vspace{-10pt}
\label{fig:extended-pbt}
\end{figure*}

\Cref{fig:single-arm-pbt,fig:dual-arm-pbt} demonstrate performance of PBT with $|\mathcal{P}|=8$ compared to regular PPO. We use equivalent amount of compute in both cases and compare the best agent in the population with the best agent among 8 independent PPO runs. We find that PBT improves training substantially in all scenarios, and in three of them PBT becomes the enabling factor allowing the algorithm to reach non-trivial performance. For example, in single-arm reorientation task, without PBT none of the eight single-GPU training sessions reached the target tolerance $\varepsilon^*$ (rightmost plot in \Cref{fig:single-arm-pbt}). Of the three types of scenarios, reorientation relies on exploration the most: finding the correct approach to in-hand manipulation is essential, and there are many local optima to get stuck in. PBT excels at overcoming these challenges. It greatly amplifies the exploration capabilities of RL by directing computational resources towards promising agents and trying multiple combinations of RL hyperparameters and reward shaping coefficients.

Our learning approach scales well even to dual-arm tasks, despite the significantly increased overall complexity and exploration challenges. Moreover, PBT dual-arm agent~(\Cref{fig:dual-arm-pbt}) performed better in the reorientation task, reaching almost 40 out of $N_{max}=50$ successes. The dual-arm task requires the agent to constantly pass the object from one hand to another, as a result the policy became more confident at juggling and tossing the object in-hand, while single-handed agents tend to rely on more conservative in-hand rotations.

To test the scaling properties of the algorithm we train populations of 16 and 32 agents on the single-arm reorientation task, allowing each agent to observe $10^{10}$ environment transitions (the total amount of collected experience in the $|\mathcal{P}|=32$ experiment is thus 0.32 trillion environment steps). We observe that each increase in population size leads to significant improvement of both convergence speed and final agent performance, reaching $N_{succ} > 42$ for the best agent (see \Cref{fig:extended-pbt}).

\Cref{fig:extended-pbt} additionally shows some hyperparameter schedules discovered by PBT. Evidently, meta-optimization tends to prefer smaller policy updates as training progresses, tightening the adaptive learning rate KL-divergence threshold (\Cref{sec:rl}). We hypothesize that high-performing policies become quite sensitive to large SGD steps, and such schedule helps enforce the trust region and prevent destructive parameter updates. 

Curiously, only our largest experiment ($|\mathcal{P}|=32$) was able to find solutions with higher $\gamma$. The highest-performing agent in this experiment employs an alternative strategy: for certain reorientations it chooses to place the object back on the table, rotate it, and then pick up again and put the object into the target position. While this approach is slower, it minimizes the risk of dropping the object and leads to better long-term outcomes. We demonstrate this and other learned behaviors in \href{https://sites.google.com/view/dexpbt}{supplementary videos}.

\section{Conclusions}
\label{sec:conclusions}

In this paper, we demonstrate the ability of end-to-end deep RL to learn control policies for sophisticated dexterous manipulators. We employ Population-Based Training at scale to train agents that are able to control high-DoF simulated robotic systems in contact-rich conditions. Although our scenarios only cover a small part of the robotic dexterity domain, we consider solving these object reposing challenges to be an important step on the path toward real-world deployment of robotic systems with human-level object manipulation capabilities.

While our agents demonstrate strong performance in simulation, many additional obstacles need to be overcome before practical applications are feasible. Our policies demonstrate aggressive control on the limits of robot capabilities which can lead to equipment damage in the real world. One promising approach is to utilize Riemannian Motion Policies~\cite{AnqiLi2021rmp2} or Geometric Fabrics~\cite{karlvanwyk2022fabrics} to improve safety on the real robot by imposing conservative motion priors.

One of the most important future research directions is closing the sim-to-real gap. One approach that has been shown to improve sim-to-real transfer is the randomization of physical parameters during training. Projects such as Dactyl~\cite{openai2020dactyl,openai2019rubiks} have demonstrated that training policies with domain randomization is particularly challenging and can require substantial computational budget. Our approach, similar to DeXtreme \cite{ahanda2022dextreme}, based on parallelized physics simulation and high-throughput GPU-accelerated learning has the potential to significantly reduce computational requirements for particularly challenging experiments involving dual arm and hand systems and make them accessible to a wider research community.

\bibliographystyle{plainnat}
\bibliography{references}

\end{document}